\def\cvprPaperID{0000}
\def\confName{CVPR}
\def\confYear{2023}

\def\paperTitle{DiffFaceSketch: High-Fidelity Face Image Synthesis with Sketch-Guided Latent Diffusion Model}

\def\authorBlock{
    Yichen Peng\dag        \qquad
    \and
    Chunqi Zhao\ddag \\    \qquad
    \and
    Haoran Xie\dag
    \and
    Tsukasa Fukusato\ddag
    \and
    Kazunori Miyata\dag   \qquad \\
    \and
    
    \dag Japan Advanced Insititute of Science and Technology \\
    \ddag The University of Tokyo \\

    {\tt\small yichen.peng@jaist.ac.jp}
}

\newif\ifreview 
\newif\ifarxiv \newcommand{\arxiv}{\arxivtrue}
\newif\ifcamera 
\newif\ifrebuttal 

\arxiv 

\pdfoutput=1
\documentclass[10pt,twocolumn,letterpaper]{article}
\ifreview \usepackage[review]{cvpr} \fi
\ifarxiv \usepackage[pagenumbers]{cvpr} \fi
\ifrebuttal \usepackage[rebuttal]{cvpr} \fi
\ifcamera \usepackage{cvpr} \fi

\usepackage{graphicx}
\usepackage{amsmath}
\usepackage{amssymb}
\usepackage{booktabs}

\usepackage{times}
\usepackage{microtype}
\usepackage{epsfig}
\usepackage[table,xcdraw]{xcolor}
\usepackage{caption}
\usepackage{float}
\usepackage{placeins}
\usepackage{color, colortbl}
\usepackage{stfloats}
\usepackage{enumitem}
\usepackage{tabularx}
\usepackage{xstring}
\usepackage{multirow}
\usepackage{xspace}
\usepackage{url}
\usepackage{subcaption}
\usepackage{xcolor}
\usepackage[hang,flushmargin]{footmisc}

\ifcamera \usepackage[accsupp]{axessibility} \fi

% ...

\ifarxiv  \fi

\newcommand{\R}[1]{{%
    \textbf{%
        \ifstrequal{#1}{1}{\textcolor{red}{R#1}}{%
        \ifstrequal{#1}{2}{\textcolor{blue}{R#1}}{%
        \ifstrequal{#1}{3}{\textcolor{magenta}{R#1}}{%
        \ifstrequal{#1}{4}{\textcolor{teal}{R#1}}{%
                           \textcolor{cyan}{R#1}%
        }}}}%
    }%
}}

\usepackage{xr-hyper}

\makeatletter
\newcommand*{\addFileDependency}[1]{
  \typeout{(#1)}
  \@addtofilelist{#1}
  \IfFileExists{#1}{}{\typeout{No file #1.}}
}

\makeatother

\usepackage[pagebackref,breaklinks,colorlinks]{hyperref}
\usepackage[capitalize]{cleveref}
\crefname{section}{Sec.}{Secs.}
\crefname{table}{Table}{Tables}
\crefname{figure}{Fig.}{Figs.}

\begin{document}
\twocolumn[{%
\title{\paperTitle}
\author{\authorBlock}

\maketitle
\begin{center}
    
  \centering
  \captionsetup{type=figure}
  \includegraphics[width=\textwidth]{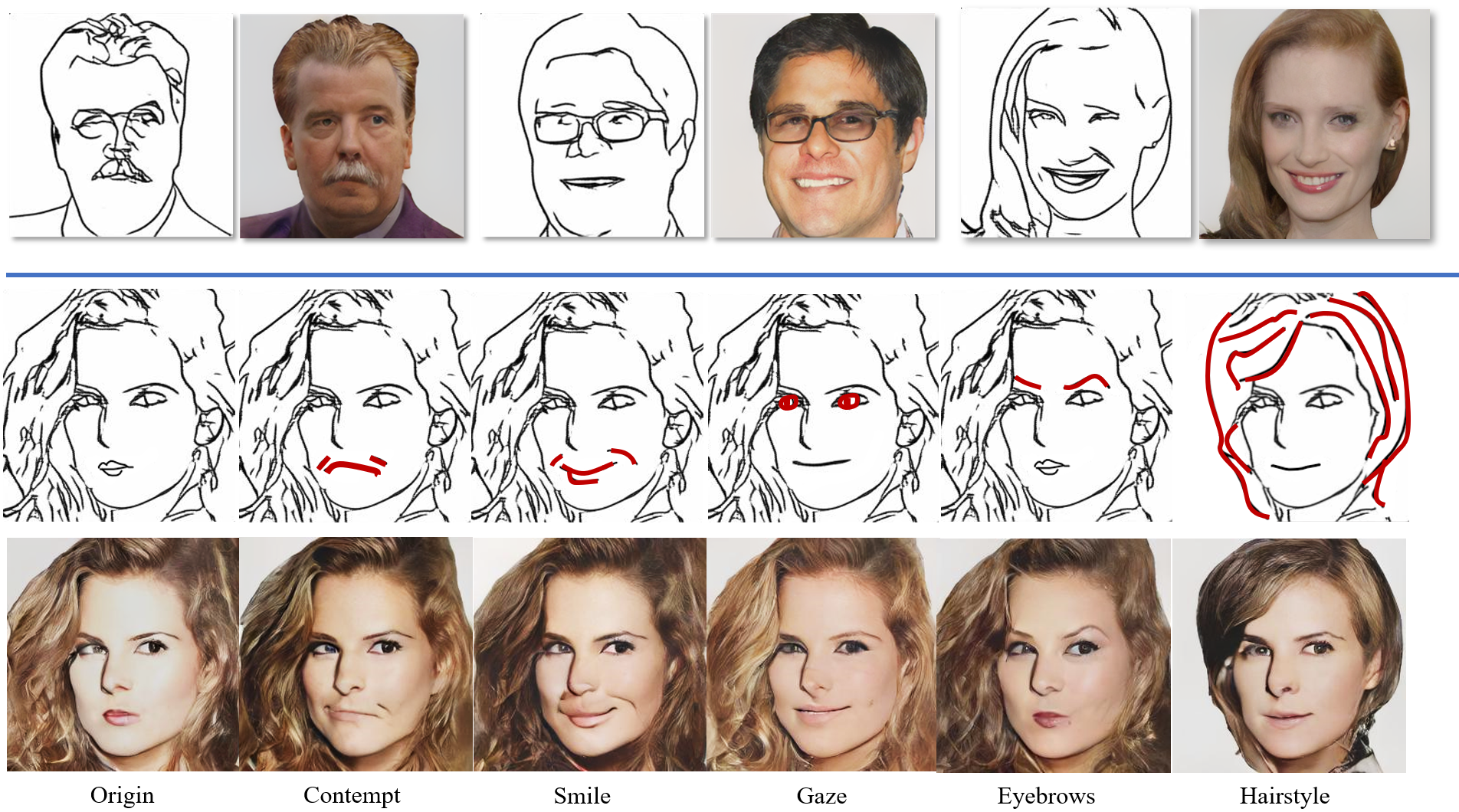}
  \caption{A Sketch-Guided Lantent Diffusion Model (SGLDM) synthesizes high-quality face images with high consistency of input sketches. 
  SGLDM enables users to simply edit face images such as different expressions, facial components, hairstyle, etc. The edited strokes are highlighted in red.}
  \label{fig:teaser}
\end{center}

}]

\begin{abstract}
Synthesizing face images from monochrome sketches is one of the most fundamental tasks in the field of image-to-image translation. 
However, it is still challenging to (1)~make models learn the high-dimensional face features such as geometry and color, and (2)~take into account the characteristics of input sketches. 
Existing methods often use sketches as indirect inputs (or as auxiliary inputs) to guide the models, resulting in the loss of sketch features or the alteration of geometry information. 
In this paper, we introduce a Sketch-Guided Latent Diffusion Model (SGLDM), a LDM-based network architect trained on the paired sketch-face dataset. We apply a Multi-Auto-Encoder (AE) to encode the different input sketches from different regions of a face from pixel space to a feature map in latent space, which enables us to reduce the dimension of the sketch input while preserving the geometry-related information of local face details. We build a sketch-face paired dataset based on the existing method that extracts the edge map from an image. We then introduce a Stochastic Region Abstraction (SRA), an approach to augment our dataset to improve the robustness of SGLDM to handle sketch input with arbitrary abstraction.
The code and dataset will be released in the project page.\url{https://puckikk1202.github.io/difffacesketch2023/}
The evaluation study shows that SGLDM can synthesize high-quality face images with different expressions, facial accessories, and hairstyles from various sketches with different abstraction levels.
\end{abstract}
\section{Introduction}
\label{sec:intro}
\begin{figure}[t]
    \centering
    \includegraphics[width=\linewidth]{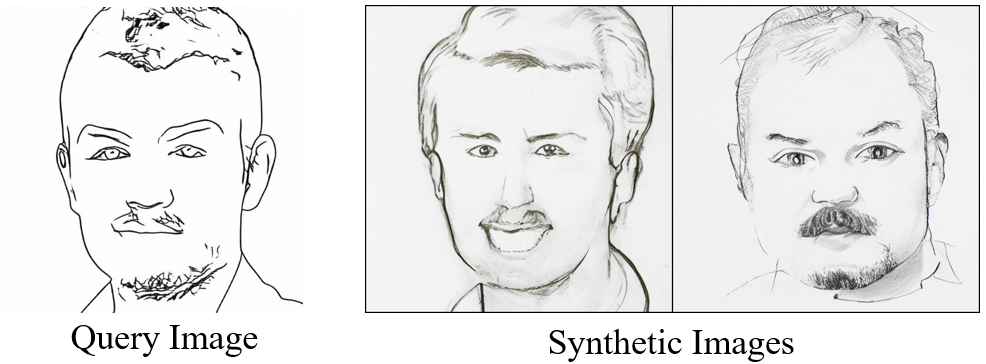}
    \caption{An example of implementing a LDM-based model, \textit{stable diffusion}~\cite{Robin2022} with pre-trained weights. Although we inputted a single sketch (left) and texts (e.g., ``\textit{a face photo}'' or ``\textit{a portrait}''), the generated results are not colored images but monochrome sketches, and do not reproduce the contours of the input sketch.}
    \label{fig:ldm_fail}
\end{figure}

Synthesizing images, especially human faces, from a monochrome sketch is one of the most fundamental tasks in the image-to-image translation manner. It benefits various applications such as character design and inmate tracking. 
However, the sparse distributions of single-channel sketch data make feature extraction and generalization difficult.
In addition, collecting paired datasets of painter's sketches and the corresponding photographs is time-consuming and labor-intensive. 
It is challenging for the synthesis model to understand the monochrome sketch input with redundant semantic information (e.g., separated facial components, expressions,accessories, and hairstyles ).
GAN-based generative models~\cite{Lee2020,Park2019} are one of the feasible solutions for sketch-to-image generation based on semantic mask annotated datasets~\cite{Lee2019,Caesar2016}. 
Although they allow users to arrange facial semantics (i.e., regional-only conditions), many details may be lost or arbitrarily synthesized, such as wrinkles and mustaches. 
Instead of applying semantic masks, the other previous GAN-based models~\cite{Chen2020} trained using sketch-face paired datasets can directly generate (and edit) face images from monochrome sketches. 
However, they are unsuitable for handling local geometrical details such as accessories and expressions since no semantic information was directly specified in rough monochrome sketches.
More recently, the diffusion model (DM)~\cite{Ho2020,song2020,alex2021} and Contrastive Language-Image Pre-training (CLIP)~\cite{Alec2021} have achieved tremendous success on the text-to-image task. 
However, in the case of image-to-image, especially sketch-to-image, their system requires not only image input but also appropriate text inputs, and may not generate desired images, as shown in \autoref{fig:ldm_fail}. 
The other conditioning-guided DM-based models such as ILVR~\cite{choi2021} and SDEdit~\cite{meng2021} approached the image-to-image task by inputting an RGB image reference to control the synthesis. However, it is generally difficult to specify image details after noise injection and resampling of the query input.
To maximize the generative models to learn from the paired sketch for more accurate information, in this work, we introduce a Sketch-Guided Latent Diffusion Model (SGLDM), an LDM-based network architect trained using a sketch-face dataset. The LDM is exceptional at flexible and high-quality inference with different conditions, we apply LDM as a backbone for our sketch-guided image synthesis training. 
We apply a Multi-Auto-Encoder~(AE) to encode query sketches from the pixel space into feature maps in the latent space of the image feature, which enables us to reduce the dimension of the sketch input while preserving the geometrical-related information of face local details. 
Moreover, we apply a 2-Stage train process to achieve a better distribution mapping between the sketch and image domains. 
Since different people focus on different facial regions, this often leads to different levels of abstraction in the input sketch. 
For example, some people focus on the details of the eyes, while others focus on the mouth. 
To access sketch data with different levels of abstraction, we introduce a data-augmentation method, named \textit{Stochastic Region Abstraction} (SRA) to improve the robustness of SGLDM, while sketch data is extracted from Celeba-HQ using sketch simplification methods~\cite{Edgar2016, Edgar2018}. The evaluation study shows our model can generate natural-looking face images from sketches with different levels of detail. In addition, SGLDM also enables users to synthesize desired face images (in the resolution of 256$\times$256) with different expressions, facial accessories, and hairstyles via a monochrome sketch (see \autoref{fig:teaser}).
In short, our main contributions are summarized as follows.
\begin{itemize}
\setlength{\leftskip}{-3mm}
    \item We proposed SGLDM, a sketch-input-only model, which trained via a 2-Stage training process to synthesize faces with high quality and input consistency. 
    \item We introduced SRA, a data augmentation strategy for synthesizing convincible faces from input sketches in different levels of abstraction. 
    \item We verified the SGLDM achieves superior scores in various metrics compared to the state-of-the-art methods and is sufficiently robust enough to generate the intended face images.
\end{itemize}

\begin{figure*}
    \centering
    \includegraphics[width=0.9\textwidth]{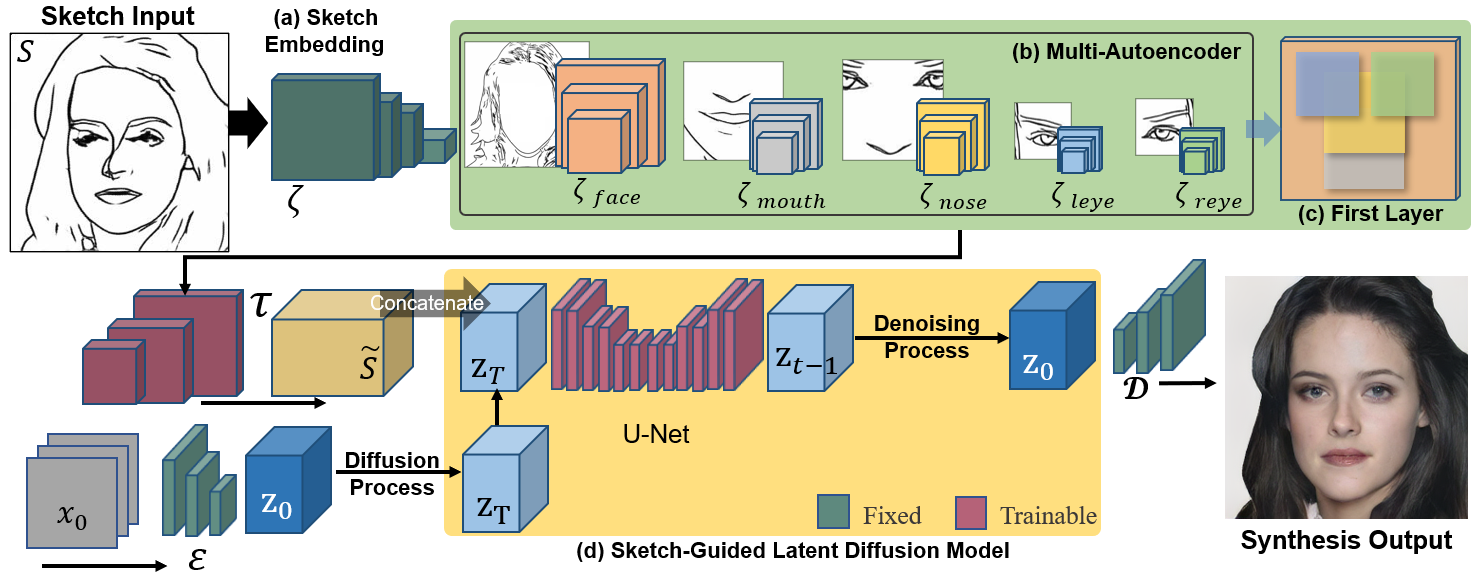}
    \caption{The framework of SGLDM. In the sketch embedding stage (a), given a sketch input $S$, a pretrained sketch encoder $\zeta$ encodes $S$ into a feature vector $S'$. The decoder $\theta$ then decoder $S'$ into a feature map $\tilde{S}$. In the latent denoising stage (b), the random latent code $Z_T$ is concatenated by the feature map $\tilde{S}$ and denoised to $Z_0$ by a U-Net. Finally, the output latent code $Z_0$ is decoded by a decoder $\mathcal{D}$ to the final output.}
    \label{fig:framework}
\end{figure*}

\section{Related Work}
\label{sec:related}
\subsection{Sketch-based Image Synthesis}
Image synthesis from sketches has been studied extensively over the last decade. From the beginning, sketch-to-image was treated as an image-to-image translation task, and researchers tried to train the deep learning-based network to eliminate the domain gap between monochrome sketches and RGB images.
Some supervised GAN-based generative models, such as Pix2Pix\cite{zhu2017} and Pix2pixHD~\cite{wang2018}, require paired sketch-image datasets constructed by extracting the edge map of real images to train the models.
To improve the performance of mapping the image domain into the sketch domain, a large number of pairwise sketches and photos are required. Then, various dataset such as Sketchycoco~\cite{gao2020}, which categorize objects into different classes, have been released. 
The facial sketch image dataset is even less accessible, such as the CUHK Face Sketches~\cite{wang2009,wang2011}.
On the other hand, unsupervised image-to-image has been introduced in some other works (e.g., CycleGAN~\cite{zhu2017}, DualGAN~\cite{yi2017}). More recently, as the disentangled representation of StyleGAN's $w+$ space has developed strongly, sketch-to-sketch is also treated as a style transfer task, for example, DualStyleGAN~\cite{yang2022} and Pixel2Style2Pixel~\cite{Elad2021}.
However, the end-to-end GAN-based models suffer from unstable training and are easily overfitted to a specific dataset, which limits the variety of synthesis results. Therefore, inspired by the recent outperformance of LDM in conditional image synthesis tasks, we propose SGLDM to enable high-quality face synthesis with high input consistency. 
\subsection{Diffusion Models}
More recently, diffusion and score-based models have flourished as powerful image synthesis models, the backbone of which is a U-Net~\cite{Olaf2015} that has achieved remarkable success in terms of diversity, quality, training stability, and module extensibility.
Previous work~\cite{Ho2020,song2020} have shown outperformance, especially in unconditional image synthesis. However, the high cost of computational resources limits the resolution of the synthesized images. Fortunately, thanks to the contribution of \cite{Robin2022}, the image is first coded from high-dimensional RGB space into a low-dimensional feature code in latent space.
The latent code is then used to process the forward and backward diffusion procedures. 
Moreover, its good modular extensibility makes it possible to deal with image-to-image tasks such as image inpainting, semantic mask-to-image, layout-to-image, etc~\cite{Robin2022}. 
Other than modifying the network architect of a DM-based model, ILVR~\cite{choi2021} and SDEdit~\cite{meng2021} proposed the conditioning-guide sampling algorithms to tackle the image-to-image task, which required a blurry RGB reference as input to iteratively guide the sampling. However, they failed to specify image detail due to the blurred conditioning.
On the other hand, for the sketch-to-image task, however, the main problem is the lack of semantic information in the monochrome sketch itself. Therefore, the sketch-to-image task via DM always requires additional input, such as a redundant text prompt. 
%
To this end, inspired by GAN-based sketch-to-image architecture~\cite{Chen2020}, we apply a \textit{Multi-Auto-Encoder} (AE) to encode the input sketch as a condition to guide the LDM denoising process. We also introduce a data augmentation method, SRA, to provide flexibility in managing sketches with different levels of abstraction.

\section{Method}
\label{sec:method}
In this section, we first give an overview of our method in \autoref{sec:overview}. We then introduce the preliminaries of DDPM and LDM in \autoref{sec:preliminaries}. The details of our framework and implementation will be discussed in \autoref{sec:SGLDM}.
\subsection{Overview}
\label{sec:overview}
\begin{figure}[t]
    \centering
    \includegraphics[width=\linewidth]{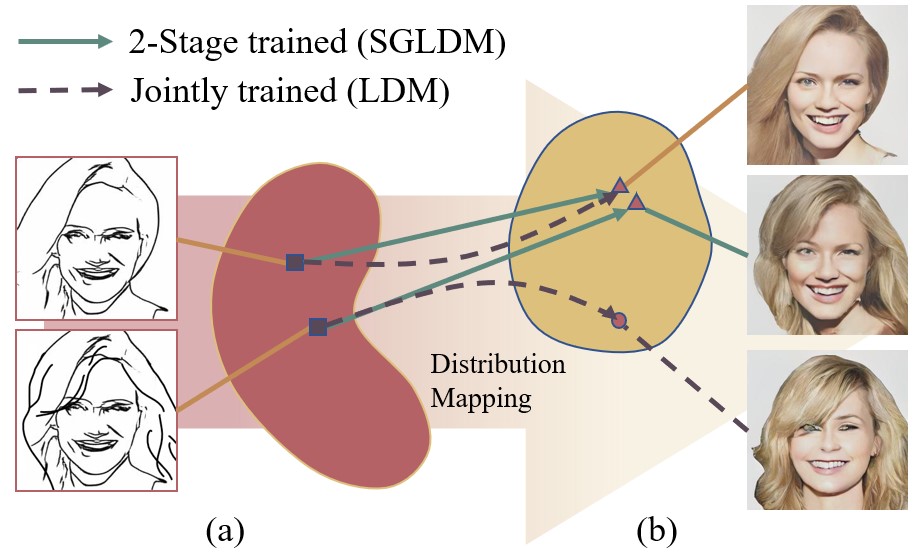}
    \caption{The different feature distribution mapping between jointly trained conditional embedding (dashed line) and the separately trained conditional embedding (solid line), from the sketch (a) to the image (b) domain.}
    \label{fig:distribution}
\end{figure}

Our goal is to synthesize a high-quality face image with high input consistency of the input sketch. We assume that the feature distribution of the monochrome sketches in the dataset is much more irregular and sparse than that of the RGB images. Jointly training a sketch embedding to map the sketch domain to the image domain may result in an unsmooth distribution (see \autoref{fig:distribution}(dashed line)). Therefore, we implemented a 2-stage training method to optimize the distribution mapping between the sketch and image domains, as shown in \autoref{fig:distribution}(solid line). More details can be found in \autoref{sec:SGLDM}. 

\subsection{Latent Diffusion Model}
\label{sec:preliminaries}
Denoising Diffusion Probabilistic Models (DDPMs or DMs) are similar to GAN and VAE, which are a new family of generative models. The key idea of DDPMs is that they randomly add noise to the initial ground true data $x_0$ gradually to distribute the data following a Markov chain to mimic the diffusion in non-equilibrium thermodynamics, which can be modeled as:
\begin{equation}
    q(x_t|x_{t-1}):=\mathcal{N}(x_t;\sqrt{1-\beta_t}x_{t-1},\beta_tI),
\end{equation}
where $t\sim[1,T]$ is a scheduled according to a pre-defined variance schedule $\{\beta_t\in(0,1)\}^T_{t=1}$ ,
They then implement and train a U-Net to learn the backward diffusion process, which can be written as:
\begin{equation}
    q(x_t|x_0) = \mathcal{N}(x_t;\sqrt{\bar{\alpha}}_tx_0,(1-\bar{\alpha}_t)I),
\end{equation}
where $\bar{\alpha}_t := \prod^t_{i=1}(1-\beta_i)$. Thus, $x_t$ can be considered as a linear combination of the initial ground true data $x_0$, and $\epsilon\sim\mathcal{N}(0,I)$. Then $x_t$ can be simplified as:
\begin{equation}
    x_t = \sqrt{\bar{\alpha}_t}x_0 + \sqrt{(1-\bar{\alpha}_t)}\epsilon
\end{equation}
The reverse process starts from $x_T$, which is close to an isotropic Gaussian $\mathcal{N}(0,\mathcal{I})$ when $T$ is theoretically large enough. The U-Net then is trained to predict $x_{t-1}$ from $x_t$ by estimating the true posterior $q(x_{t-1}|x_t)$.
This process can be modeled as:
\begin{equation}
    p_\theta(x_{t-1}|x_t)=\mathcal{N}(x_{t-1};\mu_\theta(x_t,t), \Sigma_\theta(x_t,t)).
\end{equation}
To design the loss function, Ho et al.~\cite{Ho2020} suggested simplifying the equation by re-parameterizing the Gaussian noise term to predict $\epsilon_t$ instead since $x_t$ is given in the training phase, where
\begin{equation}
    \mu(x_t,t)=\frac{1}{\sqrt{\alpha_t}}(x_t-\frac{1-\alpha_t}{\sqrt{1-\bar{\alpha}_t}}\epsilon_\theta(x_t,t))
\end{equation}
After simplification, the loss $L_{DM}$ can be written as:
\begin{equation}
    L_{DM}=\mathbb{E}_{t\sim[1,T],x_0,\epsilon_t}[\lVert\epsilon_t-\epsilon_\theta(x_t,t)\rVert^2]
\end{equation}
More recently, to reduce the computational cost, LDM has been proposed by Rombach et al.~\cite{Robin2022}. They have observed that the feature that contributes to the perceptual detail and conceptual semantic relevance remains in the latent code after being convoluted by the neural network of the AE model. Thus, a pre-trained encoder $\varepsilon$ is implemented to encode the image $x\in\mathbb{R}^{H\times W\times 3}$ in high dimensional RGB space into a lower dimensional latent code $z=\varepsilon(x)\in\mathbb{R}^{h\times w\times 3}$. Then a pre-trained decoder $\mathcal{D}$ decodes the images from the latent code $\tilde{x}=\mathcal{D}(z)$. So the loss term $Loss_{LDM}$ can be switched into:
\begin{equation}
    L_{LDM}=\mathbb{E}_{t\sim[1,T],\varepsilon(x),\epsilon_t}[\lVert\epsilon_t-\epsilon_\theta(z_t,t)\rVert^2]
\end{equation}
\section{Sketch-guided Latent Diffusion Model}
\label{sec:SGLDM}
\subsection{Framework}
Our goal is to synthesize faces following the given sketch input. To this end, we consider sketch as a condition to guide the model while denoising. \autoref{fig:framework}(a, d) shows the framework of SGLDM. Inspired by Rombach et al.~\cite{Robin2022}, we implement an LDM to lower the computational cost as well. And based on our sketch-conditioning pairs, the training loss $L_{SGLDM}$ of the conditional LDM can be written as:
\begin{equation}
    L_{SGLDM}=\mathbb{E}_{t\sim[1,T],\varepsilon(x),\epsilon_t}[\lVert\epsilon_t-\epsilon_\theta(z_t,t,\tau_\theta(\tilde{S})\rVert^2]
\end{equation}
where $\tilde{S}$ is a sketch feature encoded by a pre-trained sketch encoder $\zeta(S)$ from the input sketch $S$. And $\tau_\theta$ is a decoder to estimate a conditional map for $\varepsilon(x)$ to reverse the diffusion process. Noted that $\tau_\theta$ and $\epsilon_\theta$ are trained simultaneously.

Instead of simply jointly training a sketch encoder to provide a conditional feature map for $Z_T$ to process denoising, we construct a \textit{Conditioning Module} by pretraining a \textit{Multi-AE} network architecture. Inspired by works that separated global face into parts for the local networks such as DeepFaceDrawing~\cite{Chen2020}, APDrawGAN~\cite{YiLLR19}, and MangaGAN~\cite{su2020}, the overall encoder $\zeta$ consists of 5 partial encoders $\zeta=\{\zeta_{leye}, \zeta_{reye},\zeta_{nose}, \zeta_{mouth}, \zeta_{face}\}$, shown as \autoref{fig:framework}(b, c).

\subsection{2-Stage Training Strategy}
We adopt a 2-stage process for training. In sketch embedding stage, the \textit{Conditioning Module} is pre-trained by optimizing the sum-up MSE loss $L_{Multi-AE}$ of every single partial encoder, written as:
%
\begin{equation}
    L_{Multi-AE}=\lVert\sum_{\zeta_i\in\zeta}\zeta_i(x)-x\rVert_2
\end{equation}
%
There are 2 reasons for pre-training a \textit{Multi-AE} first rather than joint training the sketch encoders to provide a conditional feature map for the SGLDM: (1) In order for the model to learn a better mapping relationship between 2 different domain data distributions of sketch and face, so as to achieve a smoother fused domain distribution space (see \autoref{fig:distribution}). (2) To cut loads of computation cost required for training, 2-stage training separates the trainable parameters of models in each stage, thus facilitating model optimization.
We train our LDM in the second stage. To further improve the robustness of SGLDM to different sketches, we adopt the \textit{arbitrarily masking conditional training strategy} in training inspired by \textit{Masked AutoeEncoder}~\cite{He2021}, which masks random patches of the input and let the model reconstruct it automatically. In our case, since the sketch encoder $\zeta$ is pre-trained, we randomly mask out the conditioning feature map $S$ to train the denoising U-Net. 

\begin{figure}[t]
    \centering
    \includegraphics[width=1.0\linewidth]{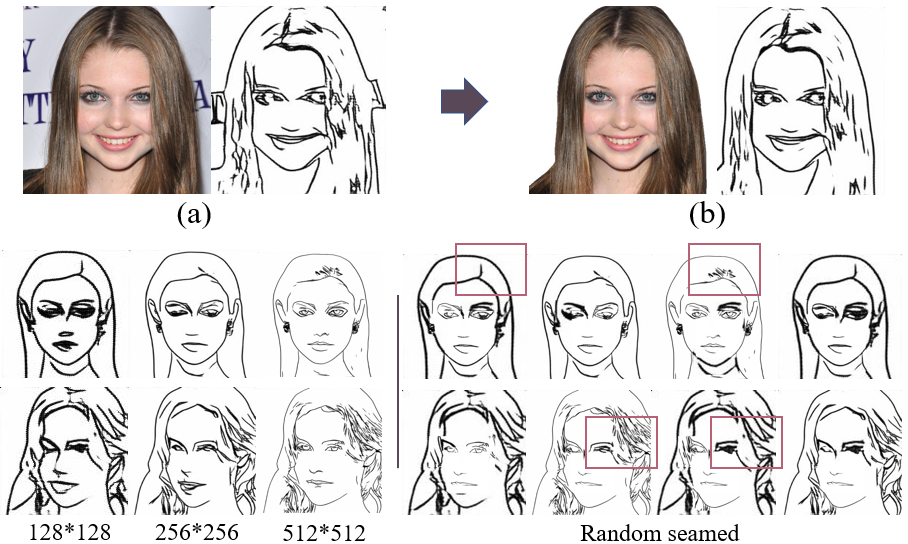}
    \caption{The original image in Celeba-HQ and its extracted edge map (a), and the result of paired data after cleaning up the background (b). Sketch simplification results from 3 different resolutions faces (left-bottom). And the random seamed data samples (right-bottom).}
    \label{fig:dataset}
\end{figure}

\subsection{Stochastic Region Abstraction Data Augmentation}
\label{sec:data}
To build our training dataset, we use high-quality 1,0000 face images from Celeba-HQ dataset~\cite{Lee2019}. We first clear up the background of the photos, as shown in \autoref{fig:dataset}(a,b). Next, we utilize \textit{sketch simplification}~\cite{Edgar2016,Edgar2018} to generate edge maps of faces. 
To enhance the robustness of SGLDM to manage sketch inputs with arbitrary abstraction, we introduce SRA to a augment the dataset.
We observed that the abstraction levels of the extracted edge maps depend on the image resolution.
Then, we resized the original photos into 128$\times$128, 256$\times$256, and 512$\times$512 resolutions respectively (see \autoref{fig:dataset}(left-bottom)), and augmented our sketch dataset. 
The red box highlighted a clear difference in different abstraction levels in the regions of hair and eyes. Moreover, following our \textit{Multi-AE} related region of every single encoder, we crop the edge maps into 5 different pieces and randomly combine them back together to a new edge map with random seams at different abstraction levels, showing as \autoref{fig:dataset}(right-bottom). We finally utilized 8k images for training, 1k for validating, and 1k for testing.
\section{Experiment and Results}
\label{sec:exp}
We conduct several experiments to verify the quality and sketch input consistency of SGLDM's synthetic face images. 

\subsection{Implementation}
Both stages of SGLDM are trained on a single NVIDIA RTX3090 GPU. In stage one \textit{Multi-AE} training, the training is performed for 500 epochs with an Adam optimizer with $\beta_1 = 0.9, \beta_2 = 0.999$, and batch size 64. The dimensions of the latent space of every AE are the same at 512. In stage two, our SGLDM is trained 300 epochs with an Adam optimizer as well, but the batch size is 8. The feature map of the sketch embedding has 8 channels, plus 3 channels of the LDM latent size, our denoising U-Net input is 11 channels of latent code and the output is 3 channels. 
\footnote{Codes, Dataset, and pre-trained models are coming soon.}
\begin{figure}[t]
    \centering
    \includegraphics[width=\linewidth]{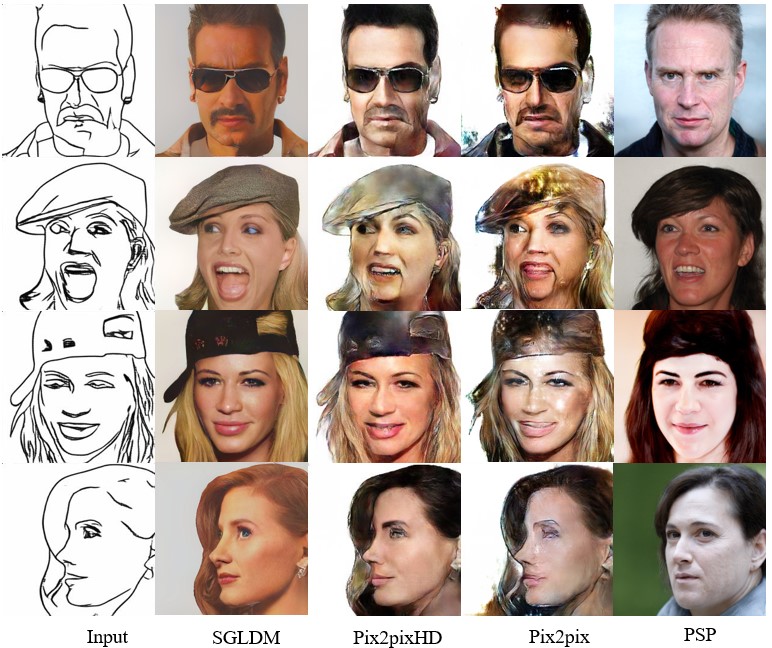}
    \caption{Examples of corner cases of sketch input which carried glasses, hat, or side face.}
    \label{fig:acc}
\end{figure}
\subsection{Quantitative Comparisons}
We compare SGLDM with several state-of-the-art image-to-image translation methods on the sketch2face task (pip2pixHD~\cite{wang2018}, pix2pix~\cite{zhu2017}, DeepFaceDrawing~\cite{Chen2020}, pixel2style2pixel (Psp)~\cite{Elad2021}, and Palette~\cite{saharia2021}). We re-trained most of the models on our 10K faces dataset picked from celeba-HQ in the same training settings. We directly implement the pre-trained weight based on 512 $\times$ 512 of DeepFaceDrawing. 

\begin{figure*}
    \centering
    \includegraphics[width=0.95\linewidth]{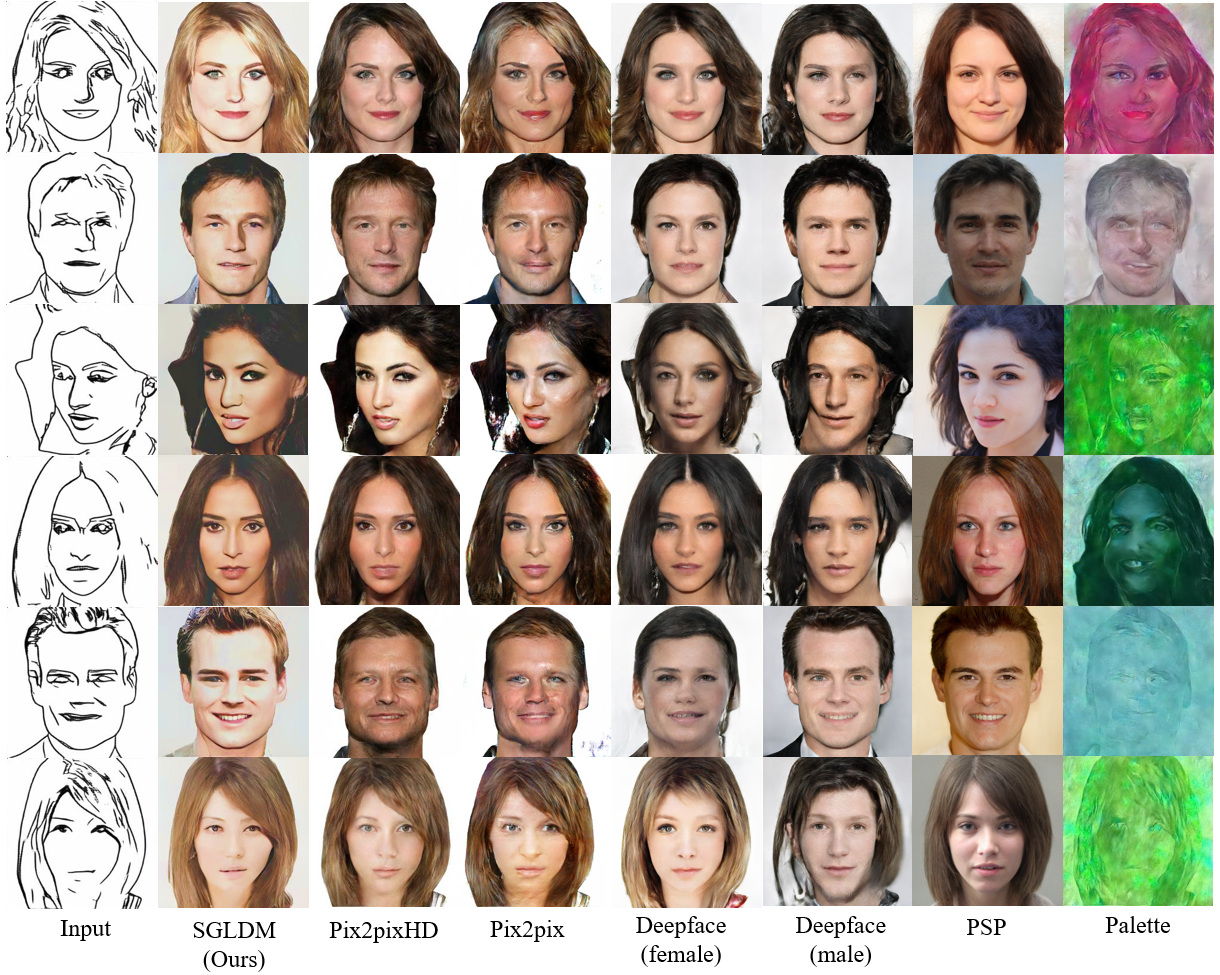}
    \caption{Qualitative comparisons of the proposed SGLDM with the state-of-the-art methods.}
    \label{fig:quatitative}
\end{figure*}

For the overall quality of results from different competing methods, as shown in \autoref{fig:quatitative}, SGLDM synthesizes more realistic faces while more faith in the input sketch. Pix2pix, Pix2pixHD and DeepFaceDrawing, however, trended to synthesize noisy faces when faces' sketches are not facing straight forward, such as the third and the last columns. Note that DeepFaceDrawing additionally required a condition to control the gender of synthetic faces, so we prepare both of the results. Although Psp achieved higher quality results visually than other methods, their methods showed poor fidelity of the sketch. 
Besides, Palette, one DM-based image-to-image method, failed to synthesize convincible faces from sketch-only input. To our knowledge, there is no state-of-the-art DM-based pipelines that rely only on monochrome sketch input, and most of them are based on text2image, segmap2image, or image inpainting pipelines fused with sketch input (e.g., \cite{horita2022}).

Next, we compared SGLDM, Pix2pixHD, and Psp which have similar fidelity results (see \autoref{fig:fidelity}). 
The black strokes on the right are input sketches to synthesize the left face images, and the red strokes behind the black strokes are filtered versions of the synthesized images using Adobe Photoshop‘s \textit{sketch filter}~\cite{adobe}. 
From the results, we confirm that SGLDM can synthesizes noiseless faces maintaining maximum consistency with the input sketch, except for some facial details such as nasolabial folds. \autoref{fig:acc} shows examples of generated face images with expressions, accessories, and hairstyles. Our method achieves a better balance between the visual quality and the consistency of inputs.

We additionally conduct a user study to compare the visual quality and the input consistency of three methods: SGLDM, Pix2pixHD, and Psp. Note that Pix2pix was not included since its visual quality is similar to Pix2pixHD. Participants were asked to choose their preferences between three types of synthetic face images generated from different models for both visual quality and input consistency. From \autoref{tab:preference}, we confirmed that face images synthesized by SGLDM achieve the highest preference for input consistency and visual quality not dissimilar to Psp.
\begin{table}[t]
    \centering
    \caption{Preference result of user study.\vspace{-3mm}}
    \label{tab:preference}
        \begin{tabular}{l|rr}
        \toprule
        Method & Quality & Fidelity \\ 
        \midrule
        Pix2pixHD~\cite{wang2018}   & 17\%    & 27\%   \\
        Psp~\cite{Elad2021}         & 37\%    & 2\%      \\ 
        Ours (SGLDM)                & \textbf{46\%}    & \textbf{71\%}   \\
        \bottomrule
        \end{tabular}
\end{table}

\vspace{1.2cm}
\begin{table*}
  \centering
  \caption{\textbf{Quantitative comparisons.} We applied the FID ($\downarrow$) score to measure the synthetic faces quality, the LPIPS ($\downarrow$) scores to evaluate the consistency between real faces and synthesized results, and a recall ratio (REC$\uparrow$) to evaluate the input consistency.\vspace{-3mm}}
  \label{tab:score}
    \begin{tabular}{l|ccc|ccc|ccc}
    \toprule
    \multirow{2}{*}{\raisebox{-5.5mm}{Method}}
    & \multicolumn{3}{c|}{Low abstraction} 
    & \multicolumn{3}{c|}{Mid abstraction} 
    & \multicolumn{3}{c}{High abstraction} \\ \cmidrule(l){2-10} 
    & FID$\downarrow$  & LPIPS$\downarrow$ & REC$\uparrow$   
    & FID$\downarrow$  & LPIPS$\downarrow$ & REC$\uparrow$ 
    & FID$\downarrow$  & LPIPS$\downarrow$ & REC$\uparrow$ \\ \midrule
    \multicolumn{1}{l|}{Pix2pix~\cite{zhu2017}}     & 53.67   & 0.20    & 0.54  & 59.46        & 0.23       & 0.50         & 63.45     & 0.28    & 0.51  \\
    \multicolumn{1}{l|}{Pix2pixHD~\cite{wang2018}}  & 51.23   & 0.18    & 0.62  & 53.71        & \textbf{0.22}       & 0.55         & 60.23     & 0.25    & 0.53  \\
    \multicolumn{1}{l|}{Psp~\cite{Elad2021}}        & 83.48   & 0.29    & 0.37  & 83.32        & 0.26       & 0.45         & 85.54     & 0.28     & 0.48  \\ \midrule
    \multicolumn{1}{l|}{SGLDM \textit{joint training}}& 46.28   & 0.20   & 0.65  & 48.62        & 0.23       & 0.54   & 50.33     & 0.26    & 0.51  \\
    \multicolumn{1}{l|}{SGLDM w/o~\textit{SRA}}  & \textbf{38.57}   & \textbf{0.17}  & \textbf{0.77}  & 48.87        & 0.26       & 0.51   & 57.76     & 0.29    & 0.48  \\
    \multicolumn{1}{l|}{Ours~(SGLDM)} & 43.58   & 0.22    & 0.71  & \textbf{45.46}    & 0.24   & \textbf{0.59}    & \textbf{46.83}  & \textbf{0.24}   & \textbf{0.57}  \\ \bottomrule
    \end{tabular}
\end{table*}
\begin{figure*}[h]
    \centering
    \includegraphics[width=\textwidth]{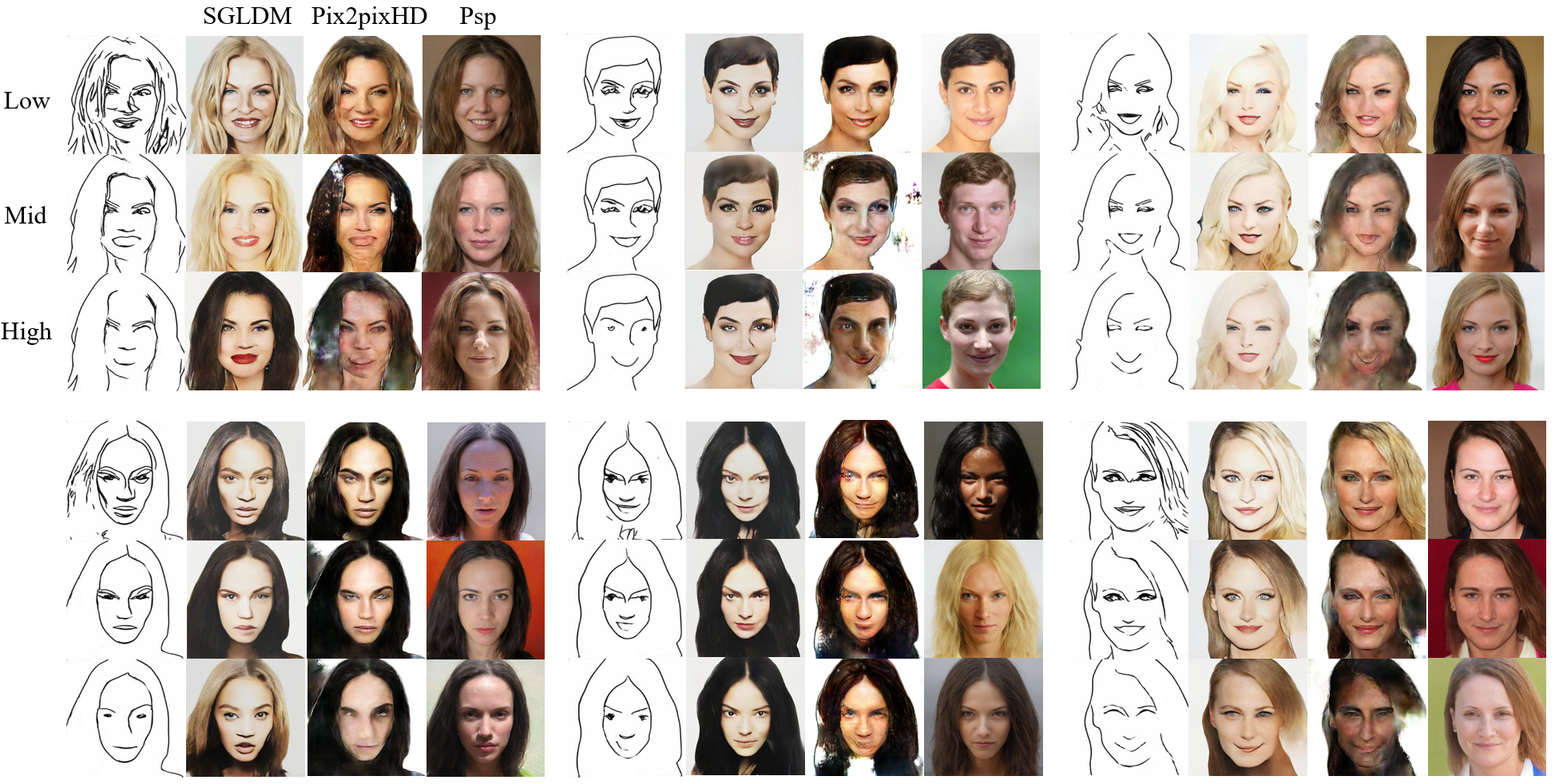}
    \caption{The comparison synthetic results in different sketch inputs with three abstraction levels.}
    \label{fig:abstraction}
\end{figure*}

\vspace{1.2cm}

\subsection{Qualitative Evaluation}
\begin{figure*}
    \centering
    \includegraphics[width=1.0\linewidth]{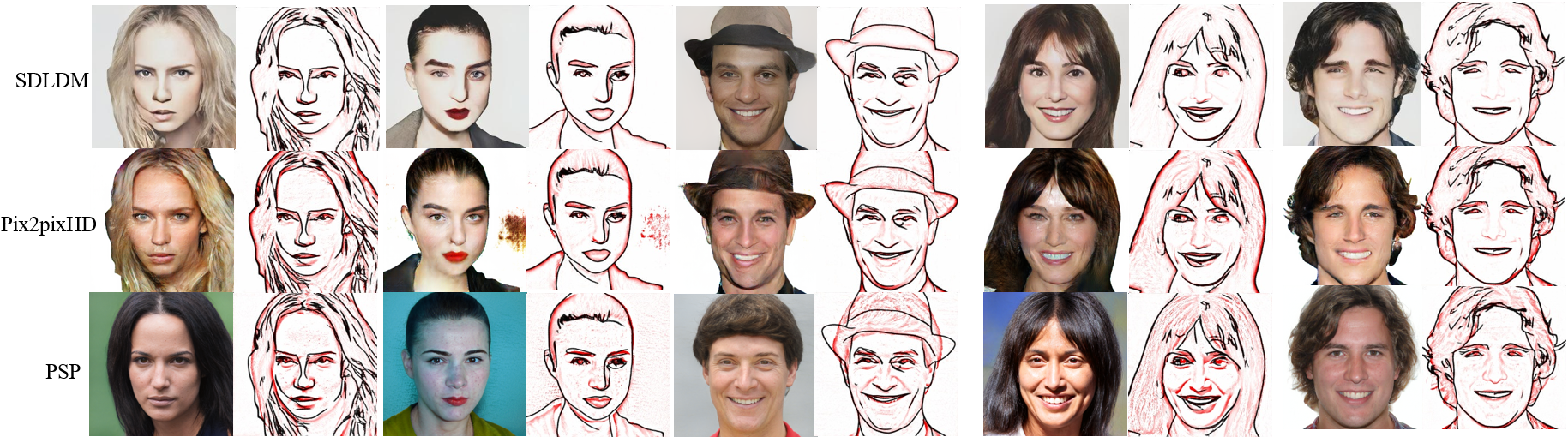}
    \caption{Fidelity comparisons of the proposed SGLDM with competing methods.} 
    \label{fig:fidelity}
\end{figure*}

For the input consistency, we calculated the recall ratio (REC) between the black and red strokes (see \autoref{fig:fidelity}). 
In addition, as the visual differences in the output are minimal with different resolutions of input sketches (as mentioned in \autoref{sec:data}), we prepared input sketches by manually erasing some strokes from the original sketches, and generated face images (see \autoref{fig:abstraction}). 
From these results, we confirmed that SGLDM is robust enough to handle rough sketches with different abstraction levels. 

\begin{figure}[ht]
    \centering
    \includegraphics[width=\linewidth]{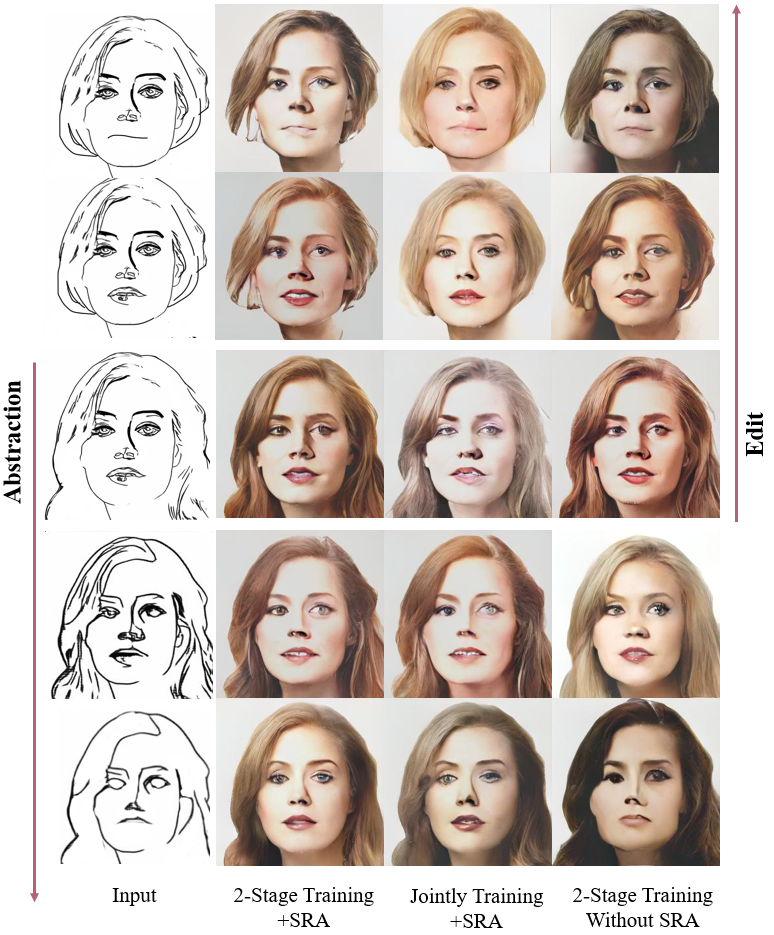}
    \caption{The synthetic faces of ablation study.}
    \label{fig:ablation}
\end{figure}

We also conducted an ablation study to compare the metrics scores between joint training \& 2-Stage training methods and to verify the validity of our SRA strategy, as shown in \autoref{tab:score}(lower-rows). 
We observed that the scores of SGLDM trained via joint training methods showed a similar performance of Pix2pixHD. 
Alghouth SGLDM trained on the single abstraction level dataset (without SRA) shows the best performance on highly detailed sketch input (low abstraction), it declined significantly under higher abstraction inputs, as shown in \autoref{fig:ablation} (rightmost column, downward). 
From the results, both separate training and data augmentation methods improved the overall performance of SGLDM on various sketch inputs (see \autoref{fig:ablation}(second column)).
\subsection{Editing Capability}
\begin{figure}[t]
    \centering
    \includegraphics[width=0.98\linewidth]{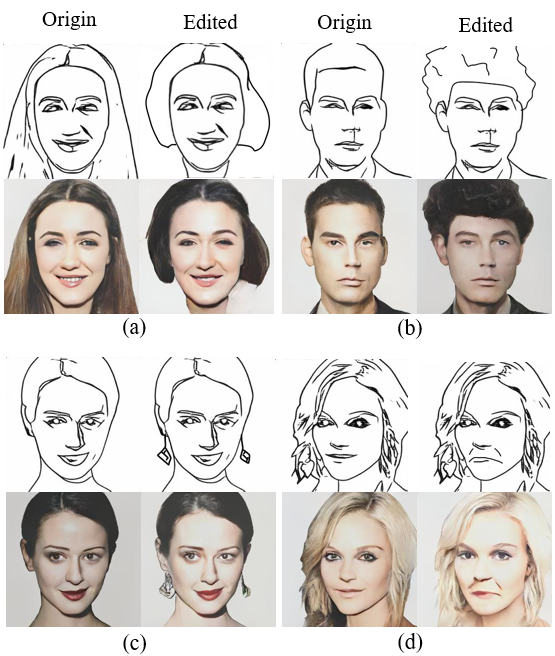}
    \caption{Examples of face editing with SGLDM. (a,b)~hairstyles, (c)~earrings, and (d)~expression.}
    \label{fig:editing}
\end{figure}

We considered the usefulness of face editing. 
\autoref{fig:editing} shows examples of partial editing (a,b)~hair style of both males and females, (c)~the earings, and (d)~expressions. 
In addition, we compared the synthetic faces of the 2-Stage trained model and the jointly trained model (see \autoref{fig:ablation}). It illustrated that the 2-stage trained SGLDM is more robust than the jointly trained SGLDM, which turns into a different identity easily after editing, see \autoref{fig:ablation}(third column, upward).
As a result, SGLDM is sufficiently robust enough to edit the intended face at will on the synthetic results.

\section{Limitations \& Future work}
\label{sec:limit}
\begin{figure}[ht]
    \centering
    \includegraphics[width=1.0\linewidth]{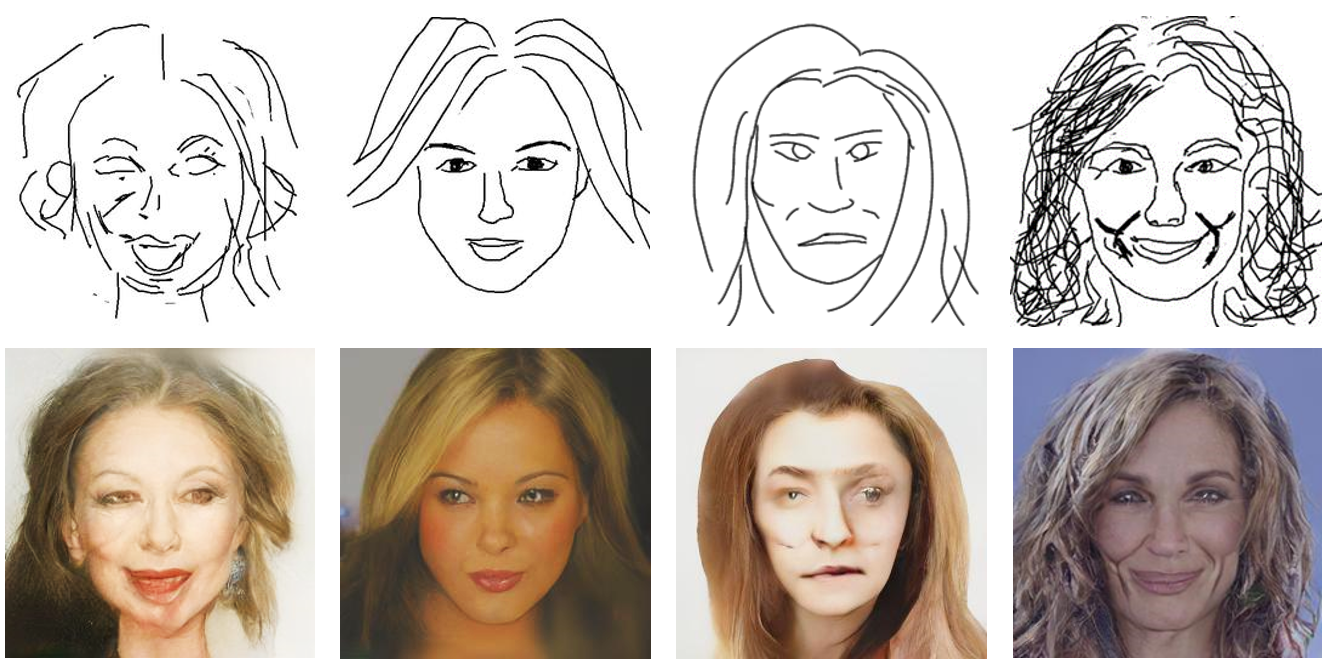}
    \caption{Less successful examples generated from the low-quality sketch inputs. Except for the second sketch from the right, the input sketches are from \cite{Chen2020}.}
    \label{fig:failure}
\end{figure}

Although SGLDM achieves high consistency with input sketches, the synthesized result tends to be too strongly affected by the input sketches. That is, noise and artifacts might be generated when inputting extremely poor sketches, as shown in \autoref{fig:failure}.
To solve this issue, some trade-off methods or algorithms will be required to keep the balance between inputs' consistency and outputs' convincibility. 
In addition to monochrome sketch input, we plan to consider a method of inputting several color cues to handle color information such as skin and hair regions. 
Moreover, we evaluated SGLDM's performance on the face synthesizing task. We believe that a similar framework can also be applied into other sketch-image tasks by changing the training dataset such as LSUN~\cite{Fisher2015} and AFHQ~\cite{choi2020}. 
And SRA can simply augmented each dataset to enhance the robustness of each model.

In this paper, although we implemented an LDM-based method to reduce the computation cost, SGLDM (i.e., the training and the sampling stage) is still computationally heavier than GAN-based models. In the training stage of a 256$\times$256 model, the maximum batch size on a single NVIDIA RTX3090 is 8, while a 512$\times$512 model's maximum batch size is only 1. In the sampling stage, the average time cost for one image is around 15.2 seconds. Although it can be cut down to around 5 to 6 seconds when using DDIM sampling strategy, the current implementation is still difficult to incorporate into a real-time interactive GUI. 
\section{Conclusion}
\label{sec:conclusion}
This paper has proposed SGLDM, a LDM-based architect face synthesizing model with a multi-AE to encode the query sketch as a conditional map while preserving the geometrical-related information of facial local details. We also introduced SRA, a data-augmentation strategy which enables the models to deal with sketch input with different abstraction levels. 
We conducted experiments to verify that SGLDM can synthesize high-quality face images with high input consistency. Moreover, SGLDM is robust enough to edit the synthetic results with different expressions, facial accessories, and hairstyles. 

{\small
\bibliographystyle{ieee_fullname}
\bibliography{11_references}
}

\ifarxiv \clearpage \appendix
\label{sec:appendix}

 \fi

\end{document}


\title{\paperTitle \\ Supplemental Material}
\author{\authorBlock}
\maketitle

\appendix
\label{sec:appendix}


{\small
\bibliographystyle{ieee_fullname}
\bibliography{11_references}
}